\documentclass[11pt,a4paper]{article}
\usepackage[hyperref]{emnlp-ijcnlp-2019}
\usepackage{times}
\usepackage{latexsym}

\usepackage{url}
\usepackage{graphicx}
\usepackage{framed}

\aclfinalcopy 

\newcommand\confname{EMNLP-IJCNLP 2019}

\graphicspath{ {images/} }

\title{Playing log(N)-Questions over Sentences}

\author{Peter Potash \and Kaheer Suleman \\
  Microsoft Research Montreal \\
  \texttt{peter.potash@microsoft.com} \\}

\date{}

\begin{document}
\maketitle
\begin{abstract}
We propose a two-agent game
wherein a questioner must be able to conjure discerning questions between
sentences, incorporate responses from an answerer, and keep track of a
hypothesis state. The questioner must be
able to understand the information required
to make its final guess,
while also
being able to reason over the game's text environment based on the answerer's
responses.
We experiment with an end-to-end model
where both agents can learn
simultaneously to play the game, showing that simultaneously achieving
high game accuracy and producing meaningful questions can be a difficult
trade-off.
\end{abstract}

\section{Introduction}
\label{sec:intro}

In recent years, the concept of reasoning
over text has drawn increased attention, primarily in the
Question-Answering community where various datasets have been developed
that are asserted to require reasoning at multiple steps
\citep{welbl2018constructing,yang2018hotpotqa}. However,
results show that the models that are the best at answering the
questions are not the best at finding the evidence humans deem important for
answering the questions\footnote{See, for example, the leaderboard for HotpotQA
\url{https://hotpotqa.github.io/} (retrieved May 21, 2019).}.
Consequently, we view it is important to \textit{modularize}
the components that supposedly conduct the reasoning in an effort to better
understand their behavior.
Furthermore, such datasets are often focused
on entities and their attributes, leaving a void in work that attempts to
perform multi-hop reasoning over the \textit{semantics} of sentences.

In an effort to fill this void, we propose a scenario where a questioner is given 
access to $N$ sentences, and the answerer chooses 1 sentence as the target
sentence to be guessed (the
answerer only has knowledge of the answer sentence, and not of the
other sentences).
The questioner has $log(N)$ yes/no questions
to determine which sentence the answerer has chosen. After $log(N)$ rounds of questions and answers, the
questioner outputs a guess of the target sentence. In order to play the game successfully,
the
questioner must output a question at each round that effectively groups the
remaining candidate sentences into 2
groups: those that do/do not possess a certain attribute.
In doing
so, the questioner must be able to form an understanding of what is similar and different among sentences in a set.
Refer to Figure \ref{fig:game example} for an example of a game with 4 candidate sentences.
With the first question, the questioner groups sentences 1 and 3 as ones that have a dog playing with an
object.
With the response of `no',
the questioner now knows the answer sentence must be either 2 or 4, and asks its second question accordingly.
The questioner is then able to use the
second response to know \textit{exactly}
the target sentence.

\begin{figure}[t]
\begin{framed}
    
\begin{footnotesize}
\begin{enumerate}
\itemsep-0.3em 
    \item a dog plays with a soccer ball
    \textbf{\item a white dog running in the backyard .}
    \item the dog is chasing a stick .
    \item the dog is sleeping .

\end{enumerate}
\begin{center}
    
\underline{Round 1 of Questioning}\\
\end{center}
Q) Is a dog playing with an object? \qquad\qquad
A) No
\begin{center}
\underline{Round 2 of Questioning}\\
\end{center}
Q) Is a dog sleeping?\qquad\qquad\qquad\qquad\qquad
A) No
\begin{center}
\underline{Prediction Round}\\
\end{center}
Q) Is it sentence 2?\qquad\qquad\qquad\qquad\qquad\,\,
A) Yes
\end{footnotesize}
\end{framed}
\caption{This is an example of two humans playing our proposed game.
The first question that
is asked must be able to generalize a property that
exists in strictly two of the sentences.
This present paper presents an initial attempt
at developing this impressive human skill.}\label{fig:game example}
\end{figure}

One can view this work as a type of generative semantic
textual similarity (STS). Related to this notion,
researchers have previously defined the task of interpretable
STS where, along with prediciton scores, models must provide alignment
between sentence parts to explain where the similarity (or lack thereof)
exists \citep{agirre2016semeval,lopez2017interpretable}. The only work
we could find that attempts to abstractly explain differences
in reference text is that of \citet{liu2018neural}, who work on the task
of automatically generating commit messages. Interestingly,
the multi-modal community has done quite similar work to ours \citep{das2017learning,Jhamtani2018learning,li2017learning},
though our work differs in that it 1)
deals solely with text; 2) requires
multiple rounds of questioning wherein responses must be noted; 3) places
an efficiency constraint on the questioner \citep{peirce1901logic}.





\section{Dataset}
\label{sec:dataset}

We create a corpus of sentence sets, aiming to group sentences with 
varying degrees of semantic relevance using
STS \citep{agirre2012semeval} and Natural 
Language
Inference (NLI) \citep{bowman2015large} datasets. We
use the following process:
1) randomly sample passage $a$ from all passages in the union of STS and
NLI; 2)
sample passage $b$ from the set of all passages
$a$ has been paired with; 3) proceed recursively (i.e. repeat
step 2 with $b$ replacing $a$) to determine passages
$c$ and $d$, making sure a given recursive path yields
four passages.
We create
a corpus with 108k sets of 4 sentences, separating
the sets into two sets: with/without
Splitting Words (SWs)\footnote{An SW is, given $N$ 
sentences, a word that appears in exactly $\frac{N}{2}$ of the sentences.
Thus, SWs only separate,
since we have sets of 4
sentences, 2 sentences from the other 2, and are therefore for the 
first round of guessing.}.
The subset \textit{with} SWs has 80k sets, while the other subset has 28k sets.
For each subset, we create splits of 80\%/10\%/10\% for 
train/dev/test. Because we do not have actual annotated game
instances, we use SWs as a heuristic for a simple
human strategy for playing the game with grounded questions.

\section{End-to-End Model}
\label{sec:e2e}

In this section we describe a model consisting of two
agents communicating over a discrete channel. This model is an important
direction because \textit{it can solve game instances that do not
contain SWs}, such as in Figure \ref{fig:game example}. Furthermore,
previous work has shown that jointly training question generation
and reading comprehension helps each task individually \citep{sachan2018self}.

\noindent\textbf{\underline{The Questioner (Q-Bot)}}
\begin{figure}[t]
    \centering
    \includegraphics[width=0.8\linewidth]{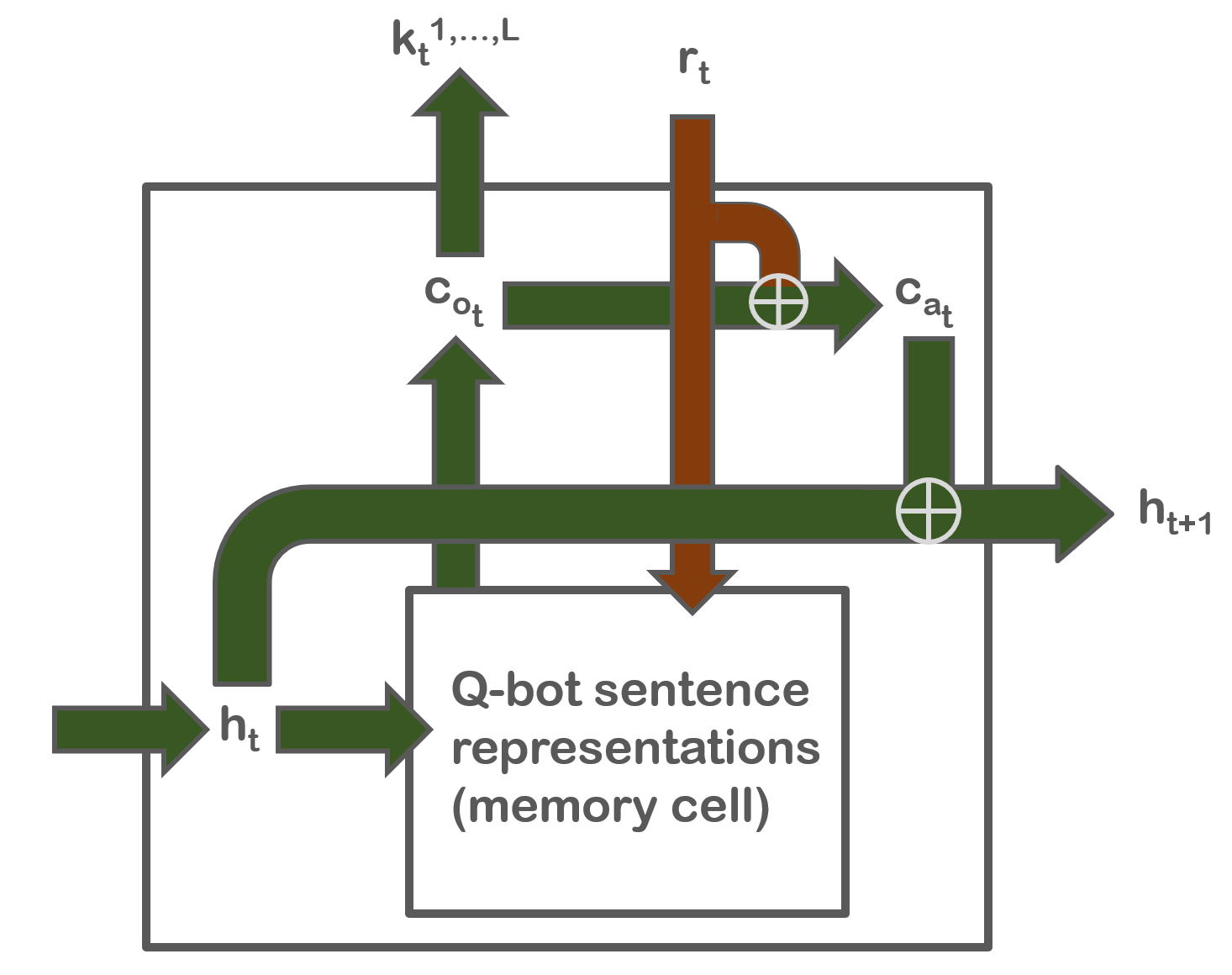}
    \caption{Inside the Q-Bot. $c_{o_t}$ is defined in Equation 
    \ref{eq:sen_com}, $k_t^{1,...,L}$ in Equation \ref{eq:query_pro},
    $c_{a_t}$ in Equation \ref{eq:combiner_ajustor}, and $h_{t+1}$ in Equation
    \ref{eq:hidden_producer}}
    \label{fig:q_bot_cell}
\end{figure}

\begin{figure}[t]
    \centering
    \includegraphics[width=0.8\linewidth]{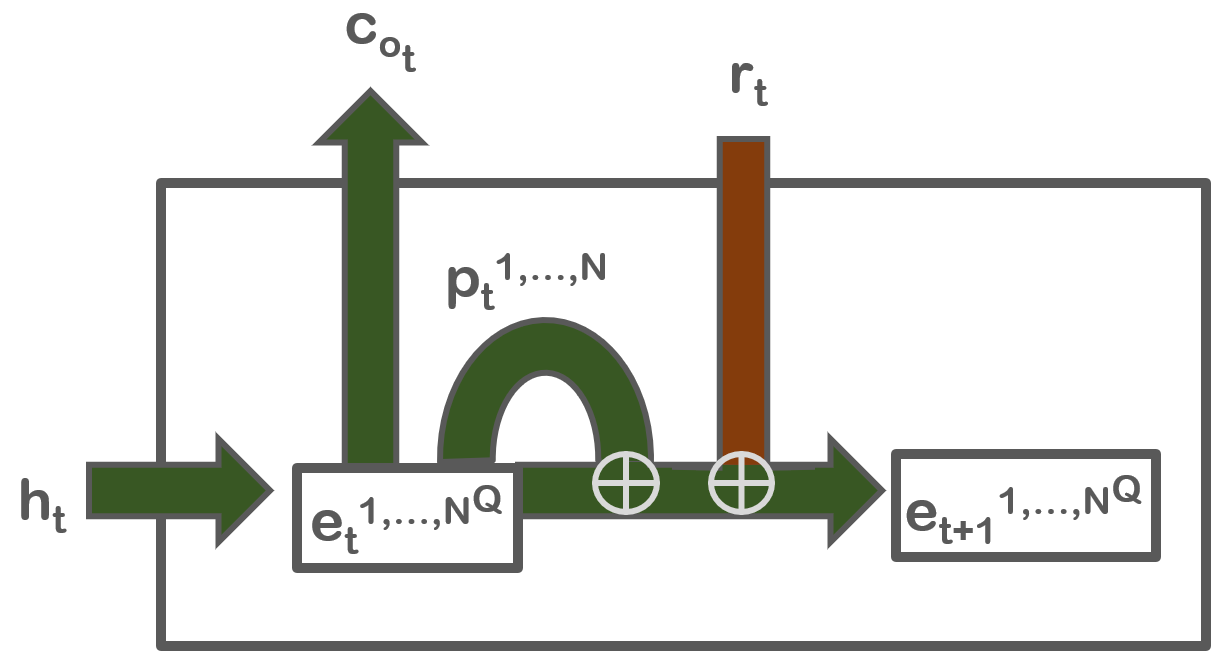}
    \caption{Inside the Q-Bot's memory cell, which holds the representations of
    the sentences at a given timestep.  $c_{o_t}, p_t^{1,...,N}$ are defined
    in Equation \ref{eq:sen_com}, and $e_t^{1,...,N^Q}$ in Equation
    \ref{eq:gater_update}.}
    \label{fig:q_bot_mem}
\end{figure}
\noindent\textbf{Sentence Encoder}: Encodes each candidate sentence individually, 
creating a real-valued
vector representation, $e_0^{i^Q}$
of each sentence, $s^{i}$, for $i
\in \{1,...,N\}$. We have chosen against an encoding
scheme that looks across the sentences at the word level because such an
approach would be unlikely to scale well across sets of many sentences.
The sentence encoding occurs once, prior to the first round of question
asking. The sentence encoder is a Bidirectional LSTM (BLSTM)
\citep{graves2005framewise}, with word embeddings initialized with GloVE
\citep{pennington2014glove}.\\
\textbf{Sentence Combiner}: Takes the individual sentence representations and
combines them to form a unified representation.
The sentence combiner also runs to produce the guess of the answer
sentence. One can consider that the this module forms a hypothesis of the
candidate answers.
Treating the sentence representations as
`memories', this module is a Memory Network \citep{sukhbaatar2015end}
(MN)
where the hidden game state $h$ attends to the sentence representations
(note $e$ is subscripted by $t$ since it is updated after each question round by Equation \ref{eq:gater_update}):
\begin{equation}
    c_{o_t}, p_t^{1,...,N} = MN(e_t^{1,...,N^Q}, h_t)\label{eq:sen_com}
\end{equation}
The probabilities used to form the weighted
sum are also used as the prediction probability over the sentences.\\
\textbf{Query Producer}: Takes the combined sentence representations and
produces a natural language question (a sequence of token
indices
with length $L$) at each 
question round using a fully-connected layer preceding a decoder LSTM.
\begin{equation}
    k_t^{1,...,L} = LSTM_{dec}^Q(\tanh(W_dc_{o_t}))\label{eq:query_pro}
\end{equation}
In order to model the fact that agents communicate over a discrete
channels, we train the Query Producer using the Gumbel Softmax 
estimator \citep{jang2017categorical}.\\
\textbf{Combiner Adjustor}: Takes the response from the answerer and adjusts
$c_{o_t}$ to create a valid hypothesis of the game
state. Since $c_{o_t}$ is used to update
$h_{t}$, if the question asked has the response `no',
the Q-Bot must act accordingly in the next round.
Given the binary response $r_t$ from the answerer,
we update $c_{o_t}$ as follows:
\begin{equation}
    c_{a_t} = r_t*c_{o_t} + (1-r_t)*\tanh(W_ac_{o_t})\label{eq:combiner_ajustor}
\end{equation}
\textbf{Sentence Gater}: Although the Combiner Adjustor updates the
`hypothesis' that directs the questioning/guessing of the Q-Bot based
on the A-Bot response, it doesn't directly increase/decrease the likelihood of
guessing a certain sentence that now seems likely/unlikely after getting a response.
Based on the response, $r_t$, we update a given sentence
representation, $e^i_t$, as follows:
\begin{equation}
    e^{i}_{t+1} = e^{i}_{t}*(w^i+\gamma)\label{eq:gater_update}
\end{equation}
\begin{equation}
    w^i = r_t*p_t^i + (1-r_t)(1-p_t^i)\label{eq:gater_probs}
\end{equation}
where $p^i$ is from
Equation
\ref{eq:sen_com}. Note that when $r=1$, $w_i$ must be 
re-normalized to create a valid probability distribution.
We add $\gamma$ to control
the relative increase/decrease of the magnitude of the sentences'
representations.
\\
\textbf{Hidden State Updater}: Uses the result of the previous round of
guessing, $c_{a_t}$, to update the hidden representation of the game state $h_t$:
\begin{equation}
    h_{t+1} = \tanh(W_hh_{t}+W_cc_{a_t}+b_h)\label{eq:hidden_producer}
\end{equation}
\textbf{\underline{The Answerer (A-Bot)}}\\
\textbf{Sentence Encoder}: Similar to the Q-Bot, but encodes,
with a BLSTM with word embeddings initialized by
GloVe,
\textit{only} the answer 
sentence, $s_i$, creating a real-valued
vector representation of the sentence, $e^{i^A}$. The A-Bot specifically has a different
encoder from the Q-Bot, to better simulate the notion that they are different
agents.
\\
\textbf{Question Encoder}: Encodes the question generated by the Q-Bot,
$k_t^{1,...,L}$,
at each
question turn using a BLSTM, creating a real-valued vector, $q_t$.
\textbf{Responder}: Takes $e^{i^A}$ with $q_t$ and produces a single bit
\textendash\, a discrete yes/no response, $r_t$, by
concatenating $e^{i^A}$ and $q_t$ together and passing it through a 2-layer
Perceptron.
The second layers produces a scalar
value, activated by sigmoid, which we convert to a discrete value via a biased straight-through
estimator \citep{bengio2013estimating}.


\subsection{Training}
\label{subsec:training}

With the gumbel softmax and straight-through estimator, we are able
to backpropogate through a differentiable estimate of the discrete
actions taken during the forward pass
and train both agents end-to-end.
The learning signal is based
on how accurate the Q-Bot is in guessing the target sentence, using the
output distribution from the sentence combiner as a soft guess of the
answer after $log(N)$ rounds of questions. The loss is
the categorical cross-entropy over all sentences, with the
target class being the answer sentence.
We also use a secondary
training signal, which is whether the Q-Bot outputs a SW
in the first question round, trained with a cross-entropy
loss over the vocabulary and the target is a SW.
A third training scenario involves pretraining
the A-Bot to correctly answer whether a question token is in the target sentence
(this setting is marked with * in Table \ref{tab:results}).

\section{Results}
\label{sec:results}

We experiment on the dataset from Section \ref{sec:dataset}\footnote{Our
code is available online *PLACEHOLDER*.}.
All training is done on the subset with SWs,
and we test on 2 different test sets:
with/without SWs (unless otherwise noted, results shown on
on the test set with SWs).
At test time, given a sentence
set, we test on 4 different game
instances, one for each sentence
as the target.
Results of our experiments are shown in Table \ref{tab:results}.
Both agents share an open vocabulary which is the first 20k tokens
of GloVe (all SWs are guarenteed to be in this vocabulary).

\begin{table}[t!]
\begin{center}
\begin{small}
\begin{tabular}{|l|c|c|}

\hline \textbf{Game Settings} & \textbf{Game Acc} & \textbf{SW Pred} \\ \hline
QL=1,loss=game & 82.2\% & 0.01\% \\
QL=5,loss=game & 84.7\% & - \\
QL=10,loss=game & 77.2\%  & - \\
QL=1,loss=sw,game & 69.8\% & 70.7\% \\
QL=1,loss=sw,game* & 74.7\% & 72.7\% \\
QL=1,loss=game,sw & 79.1\% & 49.7\% \\
QL=1,non-sw test & 80.0\% & - \\
\hline
Splitting w/ WE & - & 84\% \\ \hline

\end{tabular}
\end{small}
\end{center}
\caption{\label{tab:results} Results on
held-out data, described in Section \ref{sec:results}.
QL is the number of tokens in the question.
loss is whether the training loss is game loss alone or 
with SW grounding (the ordering dictates 
which is weighted higher).
non-sw test means the
test set used is the one \textit{without} SWs.
The final row
is described in Section \ref{subsec:split_word_emebeds}.}
\end{table}


Our results show the agents learn
to play the game effectively, trained solely with the training signal of game performance,
with performance varied slightly by question length.
However, trained with just the game signal, the agents'
strategy
does not make use of the SWs present in the game
instances.
Note that
the agents maintain similar test performance when evaluated on instances without SWs, showing that
the agents learn a strategy that is not dependant on surface forms.
By adding a secondary loss component to training, which
encourages SWs
in
the first round of questioning, the agents can learn to ask SWs, albeit
at the cost of game performance. By pre-training the A-Bot to correctly answer SW questions
(marked with * in the Table \ref{tab:results}),
we can cut the trade-off between
game performance and SW guessing: -7.5\% with,
and -12.4\% without A-Bot pre-training (comparing game accuracy
to the model trained solely on game loss). In addition, the model
increases its SW prediction by 2\%.

\subsection{Splitting with Word Embeddings}
\label{subsec:split_word_emebeds}
The final row of Table \ref{tab:results} shows how effective
pre-trained
Word Embeddings (WE)
can be used to directly determine what is the SW in a sentence set.
For all tokens in an open vocabulary, we calculate the dot product 
with all
tokens from all sentences in a given set, which, combined with the softmax function, 
produces a probability 
over the sentences.
We then find the vocab token whose
distribution minimizes the entropy between 2
sentence pairs. This result may be an upper
bound for finding SWs with an embedding based model that reflects semantic continuity,
since random embeddings, which
distill to string matching,
perfectly find splitting words.

\subsection{Qualitative Example}
\label{subsec:qual_ex}

\begin{figure}[t]
    \centering
    \includegraphics[width=0.999\linewidth]{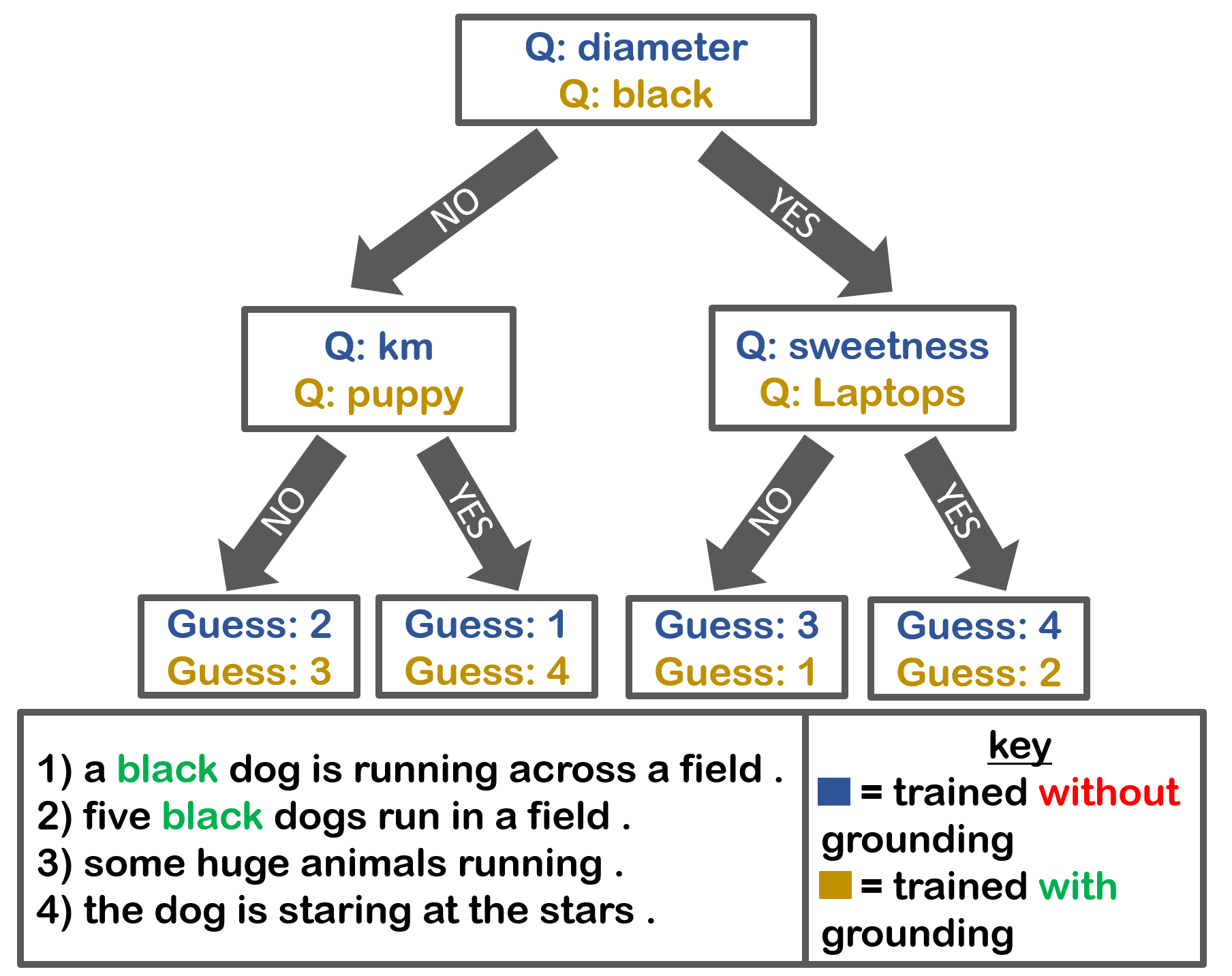}
    \caption{Given sentences at the bottom of the Figure, the results of 2
    different systems, using question
    length 1,
    arriving at their guesses over 4
    different instances of the game, one for each sentence as the target.
    Refer to Subsection \ref{subsec:qual_ex} for further discussion.}
    \label{fig:compare_example}
\end{figure}

Refer to Figure \ref{fig:compare_example} for example outputs from systems trained with/without SW
grounding. The system
outputs, with question length 1, are over 4 different game instances, differentiated by having 1 of the 4 different sentences
as the target. These are all examples of \textit{correct} guesses from each system on
each instance. Moreover, the examples show that not only do the systems use different questions to
arrive at the same guess, they also use different paths between yes/no to arrive at the same guess.
The system trained with SW grounding correctly outputs a SW as the first question.
Even more, though the second round of guessing is not grounded during training, it is able to output
a semantically meaningful token when trying to distinguish sentence 3 from 4.
Conversely, though correct, the system trained without
SW grounding lacks semantic meaning in its
questions.

\subsection{Conclusion}
\label{sec:conclusion}

We have proposed the game of $log(N)$-Questions over sentences, and introduced an end-to-end
system of 2 agents that are able to play the game. While our results show promise, there is work to be done on
improving game and SW prediction performance simultaneously, as well as playing the game over larger sentence
sets.
More generally, we shows agents exhibiting reasoning and
information-seeking in a text environment.
\bibliography{emnlp-ijcnlp-2019}
\bibliographystyle{acl_natbib}

\end{document}